\documentclass[10pt,twocolumn,letterpaper]{article}

\usepackage[pagenumbers]{cvpr}

\usepackage[accsupp]{axessibility}

\definecolor{cvprblue}{rgb}{0.21,0.49,0.74}
\usepackage[pagebackref,breaklinks,colorlinks,allcolors=cvprblue]{hyperref}
\usepackage{float}

\newcommand{\mypar}[1]{\vspace{1mm}\noindent\textbf{#1}}

\def\paperID{7992} %
\def\confName{CVPR}
\def\confYear{2025}

\title{3D-MVP: 3D Multiview Pretraining for Manipulation}

\author{Shengyi Qian$^{1,2^*}$, Kaichun Mo$^{1}$, Valts Blukis$^{1}$, David F. Fouhey$^{3}$, Dieter Fox$^{1}$, Ankit Goyal$^{1}$\\
  $^1$NVIDIA\quad $^2$University of Michigan\quad $^3$New York University  \\
  {\small \url{https://jasonqsy.github.io/3DMVP}}\\
}

\begin{document}

\maketitle
\def\thefootnote{*}
\footnotetext{The work was done during internship at NVIDIA.}

\begin{abstract}
Recent works have shown that visual pretraining on egocentric datasets using masked autoencoders (MAE) can improve generalization for downstream robotics tasks. However, these approaches pretrain only on 2D images, while many robotics applications require 3D scene understanding. In this work, we propose 3D-MVP, a novel approach for \underline{3D} \underline{M}ulti-\underline{V}iew \underline{P}retraining using masked autoencoders. We leverage Robotic View Transformer (RVT), which uses a multi-view transformer to understand the 3D scene and predict gripper pose actions. We split RVT's multi-view transformer into visual encoder and action decoder, and pretrain its visual encoder using masked autoencoding on large-scale 3D datasets such as Objaverse. We evaluate 3D-MVP on a suite of virtual robot manipulation tasks and demonstrate improved performance over baselines.
Our results suggest that 3D-aware pretraining is a promising approach to improve generalization of vision-based robotic manipulation policies.
\end{abstract}

\section{Introduction}
Building learning-based manipulation systems is challenging due to the unavailability of diverse large-scale robotics data. To address this, there has been significant interest in using computer vision techniques to learn generalizable visual representations without robotics focused data, for example by self-supervised pre-training on image datasets.
In particular, inspired by the success of masked language modeling in Natural Language Processing (NLP), several recent works have explored masked autoencoding (MAE) for visual representation learning~\cite{he2022masked,bachmann2022multimae,feichtenhofer2022masked}. MAE learns to reconstruct randomly masked patches in an input image, encouraging the model to learn high-level semantic features. When applied to egocentric videos from human demonstrations, MAE has been shown to learn representations that generalize well to downstream robotics tasks such as object manipulation~\cite{xiao2022masked,radosavovic2023real,dasari2023unbiased}.

\begin{figure}[t]
    \centering
    \includegraphics[width=\linewidth]{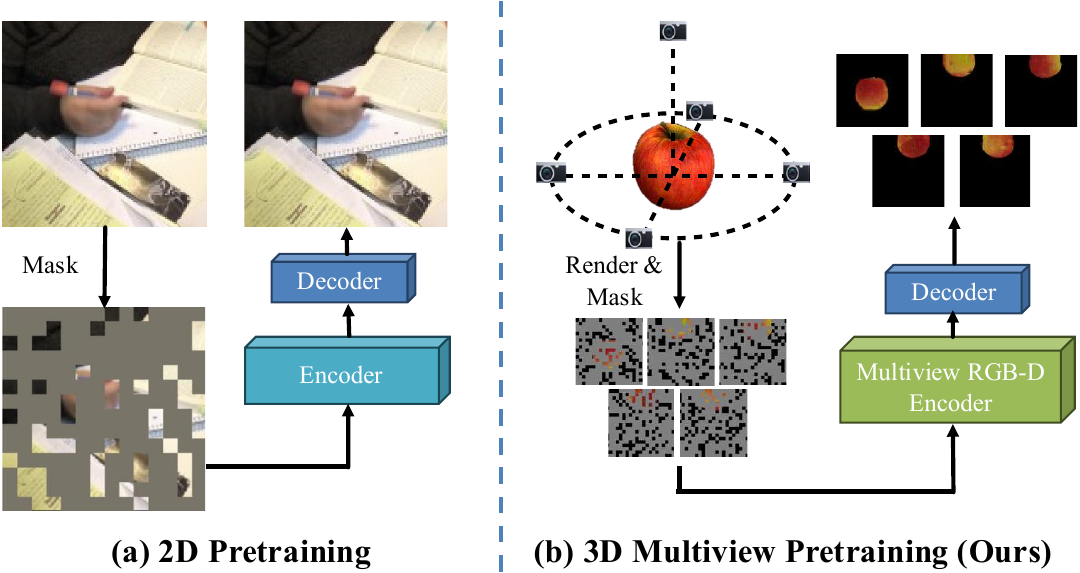}
    \caption{\textbf{Comparison of 2D vs. 3D pretraining for manipulation.}
\textbf{(Left):} In 2D pretraining~\cite{xiao2022masked}, the model is trained to do MAE reconstruction from a single image from Interaction videos~\cite{grauman2022ego4d,Shan20}.
The encoder is then used for downstream manipulation tasks.
However, the input can only be a single 2D image due to the pretraining.
\textbf{(Right):} We propose 3D Multi-View Pretraining (3D-MVP), which uses multiple orthogonal RGB-D views of a 3D model. The model is tasked with reconstructing the masked views by leveraging information across different perspectives, enabling it to learn more robust 3D spatial representations. This multi-view approach improves downstream performance in robot manipulation tasks by capturing richer scene understanding compared to 2D-only pretraining. And it is compatible with state-of-the-art 3D manipulation method such as RVT~\cite{goyal2023rvt} which takes 3D inputs.}
    \label{fig:teaser}
\end{figure}

However, current MAE approaches for robotics pretrain only on 2D images, ignoring the 3D structure of the scene. Prior works in robotics have shown that methods that build an explicit 3D visual representation of the environment are more sample efficient and generalize better than those with only 2D visual representations~\cite{goyal2023rvt,shridhar2023perceiver,pumacay2024colosseum}. Hence, in this work, we explore how we can bring the benefits of visual pretraining to robot manipulation methods that reason with explicit 3D representations.

We propose 3D-MVP, a method for 3D Multi-View Pretraining for robot manipulation. Our approach builds upon recent advances in robot manipulation. Specifically, we use the Robotic View Transformer (RVT), a state-of-the-art 3D manipulation method~\cite{goyal2023rvt}. RVT takes as input a point cloud of the scene and builds a 3D representation of the scene, using a set of fixed orthogonal ``virtual'' RGBD images. These RGBD images are fed through a transformer model that fuses information across views and predicts robot actions in the form of future gripper poses.

We choose RVT over other methods for manipulation that build a 3D representation of the scene (e.g. PerAct~\cite{shridhar2023perceiver} and Act3D~\cite{gervet2023act3d}), because other methods use either voxels or point clouds as input a transformer model, while RVT uses orthogonal RGBD images. The view-based representation makes RVT a suitable candidate for MAE pretraining.

We pretrain the multi-view representation in RVT by attaching it to a lightweight MAE decoder. We then randomly mask out a subset of the visual tokens for each view and train the model to reconstruct the multiview RGB-D images. After the pre-training, we discard the decoder. We then fine-tune the visual encoder along with RVT's action decoder on various manipulation tasks.

In order to learn generalizable and robust visual features, we use the recent works that have led to the creation of large-scale datasets of 3D scenes, such as Objaverse and 3D-FRONT~\cite{deitke2023objaverse,fu20213dfront,deitke2024objaverse,ramakrishnan2021habitat}. These datasets contain high-quality 3D scans of indoor environments along with realistic textures and materials. We use these datasets to create sets of orthogonal views that are similar to the 3D representation used in RVT. These sets of orthogonal views are then used for pretraining the visual encoder in RVT. We conduct experiments to ablate various different choices available for pre-training. Specifically, we study how masking strategies, dataset choice and dataset sizes affect the downstream manipulation performance.

Finally, we evaluate 3D-MVP on the RLBench benchmark~\cite{james2020rlbench}, a suite of manipulation tasks in a simulated environment. We find that pretraining the RVT encoder with 3D-MVP leads to significant improvements over training from scratch or pretraining with 2D MAE. These results inform how we can advance the state-of-the-art in robotic manipulation with the help of pretraining. We further evaluate 3D-MVP on the Colosseum benchmark~\cite{pumacay2024colosseum}, which tests a system's generalization across various unseen variations of manipulation tasks like object size change, color change, and lighting changes. We find that the proposed 3D-MVP method is more robust across various variations than RVT trained from scratch.

In summary, our contributions are three-fold. 

\noindent
$\bullet$ We propose 3D-MVP, a novel approach for 3D multi-view pretraining using masked autoencoding on large-scale 3D datasets.

\noindent
$\bullet$ We study how various design choices in pretraining, like masking strategy, dataset combination and sizes, affect downstream object manipulation performance. 

\noindent
$\bullet$ We demonstrate that pretraining with 3D-MVP leads to significant improvements on object manipulation tasks. We also show that 3D-MVP enables training policies that are more robust to variations such as size, texture, and lightning, on the COLOSSEUM benchmark.

We hope our work can inform future studies about pretraining for robotic applications.

\section{Related Work}

Our work builds upon several active areas of research, including self-supervised learning, visual pretraining for robotics, and learning robotic manipulation from demonstrations. 

\mypar{Self-supervised learning.}
Self-supervised learning aims to learn useful representations from unlabeled data by solving pretext tasks that do not require manual annotation. Early work in this area focused on designing pretext tasks for 2D images, such as solving jagsaw puzzles~\cite{noroozi2016unsupervised}, constrastive learning~\cite{chen2020simple,he2020momentum} or joint embedding approaches~\cite{assran2022masked,assran2023self,caron2020unsupervised,caron2021emerging,grill2020bootstrap,zhou2021ibot}. 
Most related to our work is the masked autoencoder (MAE) approach proposed by He et al.~\cite{he2022masked}, which learns to reconstruct randomly masked patches in an image. 
MAE has been shown to learn transferable representations for object detection and segmentation tasks.
Furthermore, Bachmann et al demonstrates MAE pretraining can be extended to different modalities such as semantics and depth~\cite{bachmann2022multimae}.
In this work, we extend the MAE approach to multi-view 3D scenes, enabling us to learn 3D-aware representations that are useful for robotic manipulation tasks.
Unlike MultiMAE which learns semantics and depth through direct supervision,
3D-MVP aims to learn a 3D-aware representation from multi view images.

\mypar{Visual pretraining for Robotics.}
Visual pretraining has demonstrated impressive generalization ability on computer vision tasks. Therefore, prior works have explored whether it works for robotics tasks as well. Specifically, the robotics community has trended towards learning representations using state-of-the-art self-supervised vision algorithms on diverse interaction datasets~\cite{grauman2022ego4d,Shan20,Damen18}, and finetune the network on robotics tasks~\cite{ma2022vip,nair2022r3m,xiao2022masked,radosavovic2023real,majumdar2024we,dasari2023unbiased,seo2023multi}.
3D-MVP follows the same procedure.
However, existing robotics pretraining approaches typically learn a 2D visual encoder (e.g. ResNet~\cite{he2016deep} or ViT~\cite{dosovitskiy2020vit}), we find they are inferior than manipulation policies which do explicit 3D modeling (e.g. RVT~\cite{goyal2023rvt}, Act3D~\cite{gervet2023act3d}).
Migrating a pretrained ViT to 3D manipulation policies is nontrivial since they do not have a 2D visual encoder.
In this paper, we propose 3D-MVP, which does 3D-aware pretraining on 3D manipulation policies, to fill the gap.

\mypar{Learning manipulation from demonstrations.}
Recent work has explored using transformers for multi-task manipulation policies that predict robot actions from visual and language inputs~\cite{shridhar2023perceiver,guhur2023instruction,liu2022instruction,shafiullah2022behavior,simeonov2023shelving}. End-to-end models like RT-1~\cite{brohan2022rt1}, GATO~\cite{reed2022generalist}, and InstructRL~\cite{liu2022instruction} directly predict 6-DoF end-effector poses but require many demonstrations to learn spatial reasoning and generalize to new scenes. 
To better handle 3D scenes, PerAct~\cite{shridhar2023perceiver} and C2F-ARM~\cite{james2022coarse} voxelize the workspace and detect the 3D voxel containing the next end-effector pose. However, precise pose prediction requires high-resolution voxels which are computationally expensive.
Recently, RVT~\cite{goyal2023rvt} proposes a multi-view transformer that attends over point cloud features from multiple camera views to predict actions. This avoids explicit voxelization and enables faster training and inference than PerAct. Act3D~\cite{gervet2023act3d} represents the scene as a continuous 3D feature field and samples points to featurize with attention, allowing adaptive resolution. GNFactor~\cite{ze2023gnfactor} jointly optimizes a generalizable neural field for reconstruction and a Perceiver for decision-making.
In contrast, our proposed 3D-MVP learns 3D scene representations through masked autoencoding pretraining on a large dataset of 3D object models. This pretraining enables 3D-MVP to build a rich understanding of 3D geometry and semantics prior to finetuning on downstream manipulation tasks. 
Compared to RVT and Act3D which train from scratch on target tasks, 3D-MVP's pretraining leads to improved performance, sample efficiency and generalization.
Unlike GNFactor which relies on a pretrained VLM to inject semantics, 3D-MVP directly learns 3D semantic features from object models.

\section{Approach}

In this section we first provide essential background on RVT, then define our method 3D-MVP that learns 3D-aware representations for robotic manipulation using masked autoencoding on multi-view 3D scenes, and finally describe how we finetune the method on downstream manipulation tasks. Figure~\ref{fig:approach} gives an overview of our approach.

\subsection{Background on Robotic View Transformer (RVT).}
\label{sec:rvt}

RVT is a state-of-the-art object manipulation method~\cite{goyal2023rvt}. It creates an explicit 3D representation of the scene by using orthogonal virtual views of a scene. Please refer to Goyal et al.~\cite{goyal2023rvt} for a full explanation. Here, we provide a brief overview and define the notation.

\begin{figure*}[t]
    \centering
    \includegraphics[width=0.95\linewidth]{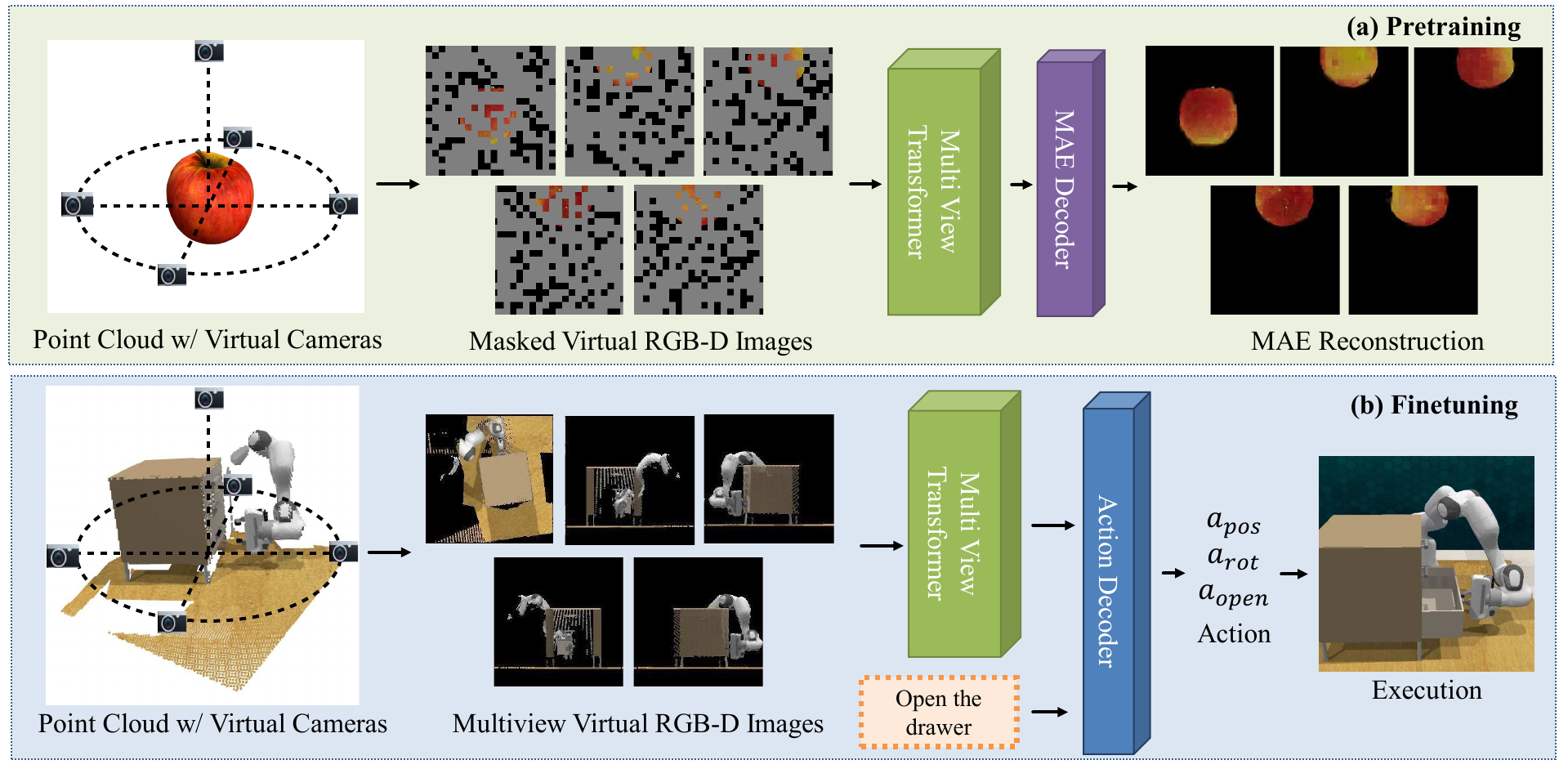}
    \caption{Overview of 3D-MVP. (a) We first pretrain a Multiview 3D Transformer using masked autoencoder on multiview RGB-D images. (b) We then finetune the pretrained Multiview 3D Transformer on manipulation tasks. Since the MVT is pretrained, the learned manipulation policy generalizes better. For example, it is more robust to changes of texture, size and lighting.
    }
    \label{fig:approach}
\end{figure*}

RVT takes a point cloud of the robot workspace as input (Fig.~\ref{fig:approach}). RVT is agnostic to the poses of the RGB-D cameras used to construct the input point cloud. For example, it can be obtained from a combination of third-person cameras around the workspace, head cameras, or wrist cameras. RVT then renders this point cloud using a set of five ``virtual" cameras placed at orthogonal locations around the robot. The virtual cameras are placed at the top, left, right, front, and back of the robot workspace with respect to the robot. Each virtual image has 10 channels: RGB (3 channels), Depth (1 channel), 3D point coordinate in world frame (3 channels), and 3D point coordinate in camera sensor frame (3 channels). We denote the virtual images captured from different virtual camera poses $\{p_1, \ldots, p_5\}$ as $\{I_1, \ldots, I_5\}$. 

These virtual images are then tokenized into $N$ patch embeddings~\cite{dosovitskiy2020vit}, flattened into a sequence of $5N$ tokens spanning all images, and fed to a multi-view transformer. The goal of RVT's multi-view transformer is to learn a function $f_\theta$ that maps the virtual images as well as language instructions $L$ to the 6-DoF end-effector pose and the gripper's binary open or close state:
\begin{equation}
    a_{\mathrm{pos}}, a_{\mathrm{rot}}, a_{\mathrm{open}} = f_\theta(L, I_1, p_1, \ldots, I_5, p_5)
\end{equation}

RVT is trained end-to-end from scratch on sampled trajectories from simulator or real robots. While RVT has shown state-of-the-art results on 3D manipulation, it does not generalize to novel objects and scenes, since we train RVT from scratch and it overfits on training demonstrations.
In the next section, we describe our novel approach 3D-MVP, and how we modify and pretrain the RVT encoder using 3D-MVP, to learn a generalizable representation.

\subsection{3D Multi-View Pretraining (3D-MVP)}
\label{sec:pretraining}

\mypar{Architecture change to RVT.}
The key idea is to pretrain the RVT visual encoder $f_\theta$ using masked autoencoding on large-scale 3D scene datasets. 
However, RVT's multiview transformer $f_\theta$ is an end-to-end model that takes language instructions as input, and produces robot actions. Existing robotics data with language and actions is limited in terms of diversity of 3D scenes, and 3D scene datasets do not typically contain robotics annotations.

To enable pre-training on 3D scene datasets, we first split the multiview transformer $f_\theta$ into an input renderer $\mathcal{R}$, an encoder network $\mathcal{E}$ and an action decoder network $\mathcal{D}$.
The renderer $\mathcal{R}$ maps the posed input images into the five virtual images, by constructing a point cloud and rendering it from the five views: 
\begin{equation}
\{I_1, \ldots, I_5\} = \mathcal{R}(I_1, p_1, \ldots, I_5, p_5)
\end{equation}
The encoder $\mathcal{E}$ maps the virtual images into a latent embedding $z \in \mathbb{R}^{5N\times H}$ (where $H$ is the hidden size) and the action decoder $\mathcal{D}$ maps $z$ to the robotic action space, i.e.,
\begin{equation}
    a_{\mathrm{pos}}, a_{\mathrm{rot}}, a_{\mathrm{open}} = \mathcal{D}(L, z), \quad z = \mathcal{E}({I_1, \ldots, I_5})\;,
\end{equation}
where tokenization of the virtual images into $5N$ patch embeddings is subsumed into $\mathcal{E}$. 
Both encoder $\mathcal{E}$ and decoder $\mathcal{D}$ are multiview transformers. 
We keep the decoder lightweight to focus on pretraining of the encoder.

\mypar{Pretraining encoder $\mathcal{E}$.}
Our visual pretraining focuses on learning a generalizable representation for the encoder $\mathcal{E}$.
We extract point clouds from Objaverse and render the point cloud using the same five ``virtual'' cameras.
Given 5 virtual images $\{I_1, \ldots, I_5\}$, we randomly mask out a subset of the visual tokens for each view, and denote the masked inputs as $\{I_1', \ldots, I_5'\}$.
We use the encoder to extract the embedding $z$ from the masked inputs,
\begin{equation}
    z = \mathcal{E}(\{I_1', \ldots, I_5'\})
\end{equation}
We use a separate, lightweight MAE decoder $\mathcal{D}_{MAE}$ to reconstruct the original image $\{I_1, \ldots, I_5\}$ from the embedding $z$.
\begin{equation}
    \{\tilde{I_1}, \ldots, \tilde{I}_5\} = \mathcal{D}_{MAE}(z)
\end{equation}
The encoder $\mathcal{E}$ and decoder $\mathcal{D}_{MAE}$ are trained end-to-end using a pixel-wise reconstruction loss:
\begin{equation}
    \mathcal{L}_{\text{recon}} = \frac{1}{5WH}\sum_{i=1}^5\sum_{p=1}^{W\cdot H} \|[I_i]_{(p)} - [\tilde{I}_i]_{(p)}\|_2^2 \; ,
\end{equation}
where $[I]_{(p)}$ indexes the image $I \in \mathbb{R}^{W \times H \times C}$ at pixel $p$.
By jointly learning to reconstruct all five images and varying the masking patterns during training, we hypothesize that the encoder will learn to reason across the multiple views and extract 3D-aware features that are robust to occlusions and viewpoint changes. In order to inform future works, we study how various masking strategies and dataset combinations affect the final downstream performance (See Tab.~\ref{tab:ablation}).

\mypar{Implementation details.}
We implement the pretraining using the PyTorch library and train it on 8 NVIDIA V100s. 
We use the Objaverse dataset for pretraining~\cite{deitke2023objaverse}, which contain a total of 800K+ 3D objects with realistic textures and materials.
We sample 200K high-quality 3D models.
and 1000 for validation purpose.
We do not construct a large-scale validation set, since the validation is only qualitative.
The evaluation of pretrained representation should be mainly done on downstream manipulation tasks.

We use a patch size of 10x10 to tokenize images.
For the encoder $\mathcal{E}$, we use a multiview transformer with 8 layers, 8 attention heads and a hidden dimension of 1024. For the decoder, we use a multiview transformer with 2 layers and 8 attention heads. We train the model for 15 epochs using the AdamW optimizer with a learning rate of 0.0001 and a weight decay of 0.01. We use a batch size of 3 and a masking probability of 0.75.

\begin{figure*}[t]
    \centering
    \includegraphics[width=\linewidth]{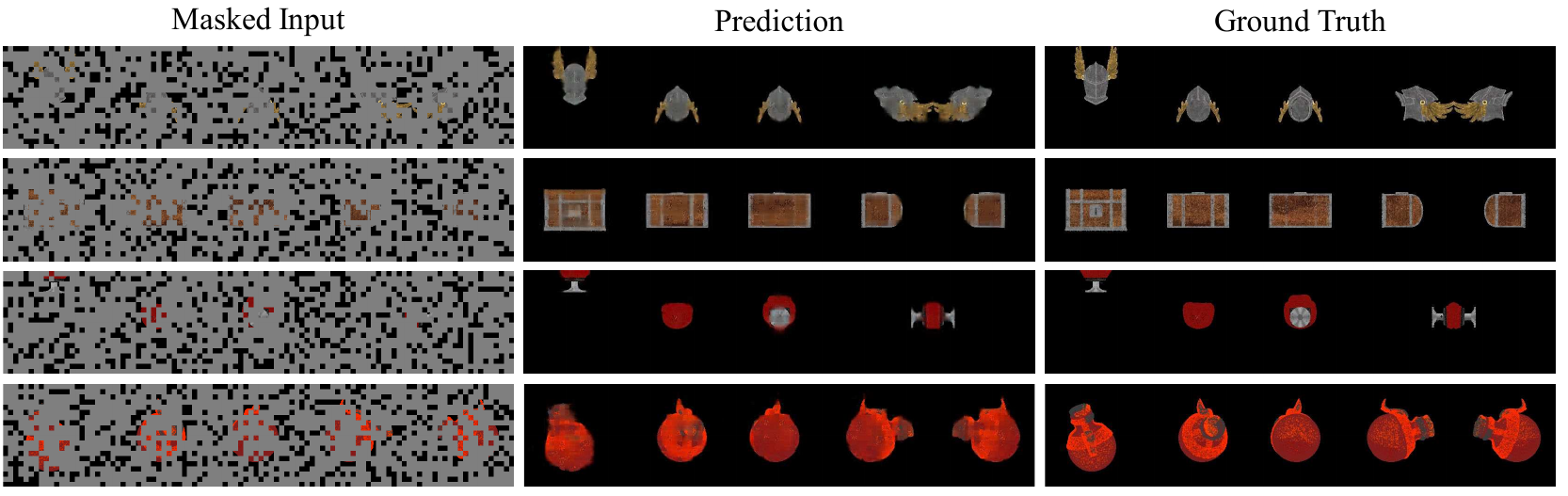}
    \caption{MAE Reconstruction results on Objaverse. Our pretrained multi-view transformer generalizes to unseen object instances and reconstructs multi-view images from their masked versions.}
    \label{fig:mae}
\end{figure*}

\subsection{Finetuning on Downstream Manipulation Tasks}
\label{sec:finetuning}

To adapt the pretrained encoder $\mathcal{E}$ for a specific manipulation task, we finetune it along with the action decoder $\mathcal{D}$ on a dataset of manipulation demonstrations.
Assume a demonstration consists of tuples of virtual images, language goals and actions.
During training, we randomly sample a tuple and supervise the network to predict the action given the virtual images and the language goal.

\mypar{Implementation details.}
For finetuning on manipulation demonstrations, we follow the standard practice~\cite{goyal2023rvt,pumacay2024colosseum} to ensure fair comparison with baselines with 2D pretraining and without pretraining.
We use 8 NVIDIA V100 (32GB) for finetuning and a single V100 for evaluation.
The learning rate is 1e-4 with 2000 warmup steps and cosine learning rate decay until 1e-6.
We use Lamb~\cite{you2019lamb} as the optimizer and the batch size is 3.
We train the model for 15 epochs on both RLBench and COLOSSEUM training set.

\section{Experiments}
\label{sec:experiments}

In this section, we evaluate the effectiveness of 3D-MVP for robotic manipulation tasks, and aim to answer the following questions:
(1) Does 3D-aware pretraining improve manipulation performance compared to training from scratch or 2D pretraining?
(2) Does 3D-aware pretraining improve robustness to environmental variations encountered in manipulation tasks?
(3) How do various design choices while pretraining affect the downstream manipulation performance?
To answer these questions, we evaluate 3D-MVP on two benchmarks: RLBench~\cite{james2020rlbench} for general manipulation performance and COLOSSEUM~\cite{pumacay2024colosseum} for systematic evaluation of robustness to environmental perturbations.

\subsection{Validating 3D Masked Autoencoding}
We validate whether masked autoencoder works in our setup with multi-view images from 3D assets. Specifically, we check whether the pretrained multi-view transformer generalizes to unseen 3D assets from Objaverse~\cite{deitke2023objaverse}. We validate it qualitatively in Figure~\ref{fig:mae}. We find that the pretrained 3D-MVP network achieves high-fidelity reconstructions despite 75\% of the input points being masked, suggesting that 3D-MVP learns meaningful 3D representations through this pretext task. 

\begin{table*}[t]
\centering
\resizebox{\textwidth}{!}{

\begin{tabular}{lccccccccccccccccc} 
& \multicolumn{2}{c}{Average}                 & Close & Drag & Insert & Meat off & Open   & Place & Place & Push  \\

Models                              & \multicolumn{2}{c}{Success} & Jar   & Stick & Peg   & Grill    & Drawer & Cups  & Wine & Buttons \\
\toprule
Image-BC (CNN)~\cite{jang2022bc,shridhar2023perceiver} & \multicolumn{2}{c}{1.3} & 0.0 & 0.0 & 0.0 & 0.0 & 4.0 & 0.0 & 0.0 & 0.0\\
Image-BC (ViT)~\cite{jang2022bc,shridhar2023perceiver} & \multicolumn{2}{c}{1.3} & 0.0 & 0.0 & 0.0 & 0.0 & 0.0 & 0.0 & 0.0 & 0.0 \\
C2F-ARM-BC~\cite{james2022coarse,shridhar2023perceiver} & \multicolumn{2}{c}{20.1} & 24.0 & 24.0 & 4.0 & 20.0 & 20.0 & 0.0 & 8.0 & 72.0\\
PerAct~\cite{shridhar2023perceiver}      &  \multicolumn{2}{c}{49.4} & 55.2 & 89.6 & 5.6 & 70.4 & \textbf{88.0} & 2.4 & 44.8 & 92.8 \\
RVT~\cite{goyal2023rvt}  & \multicolumn{2}{c}{62.9} & 52.0   & 99.2  & 11.2 & 88.0 & 71.2 & \textbf{4.0} & 91.0 & \textbf{100}\\
\textbf{3D-MVP (Ours)} & \multicolumn{2}{c}{\textbf{67.5}} & \textbf{76.0} & \textbf{100} & \textbf{20.0} & \textbf{96.0} & 84.0 & \textbf{4.0} & \textbf{100} & 96.0 \\
\bottomrule
 & Put in &  Put in & Put in & Screw & Slide & Sort & Stack & Stack & Sweep to & Turn \\
Models                               & Cupboard & Drawer & Safe & Bulb & Block & Shape & Blocks & Cups & Dustpan & Tap \\
\toprule
Image-BC (CNN)~\cite{jang2022bc,shridhar2023perceiver} & 0.0 & 8.0 & 4.0 & 0.0 & 0.0 & 0.0 & 0.0 & 0.0 & 0.0 & 8.0\\
Image-BC (ViT)~\cite{jang2022bc,shridhar2023perceiver} & 0.0 & 0.0 & 0.0 & 0.0 & 0.0 & 0.0 & 0.0 & 0.0 & 0.0 & 16.0 \\
C2F-ARM-BC~\cite{james2022coarse,shridhar2023perceiver} & 0.0 & 4.0 & 12.0 & 8.0 & 16.0 & 8.0 & 0.0 & 0.0 & 0.0 & 68.0\\
PerAct~\cite{shridhar2023perceiver}  & 28.0 & 51.2 & 84.0 & 17.6 & 74.0 & 16.8 & 26.4 & 2.4 & 52.0 & 88.0 \\
RVT~\cite{goyal2023rvt}  & 49.6 & 88.0 & 91.2 & 48.0 & \textbf{81.6} & \textbf{36.0} & 28.8 & 26.4 & 72.0 & 93.6 \\
\textbf{3D-MVP (Ours)} & \textbf{60.0} & \textbf{100.0} & \textbf{92.0} &  \textbf{60.0} & 48.0 & 28.0 & \textbf{40.0} & \textbf{36.0} & \textbf{80.0} & \textbf{96.0} \\
\bottomrule
\end{tabular}
}
\caption{Results on RLBench~\cite{james2020rlbench}. We report the task completion success rate for 18 RLBench tasks, as well as the average success rate. 3D-MVP reaches the state-of-the-art performance on the benchmark. The pretraining is mainly helpful for tasks with medium difficulty.}
\label{tab:rlbench}
\end{table*}

\subsection{Results on RLBench}
\label{sec:rlbench}
We then evaluate whether our proposed pretraining improves manipulation performance.
The experiments are conducted RLBench, a simulated platform~\cite{james2020rlbench}.

\mypar{Setup.}
RLBench~\cite{james2020rlbench} is a
popular simulation benchmark for learning manipulation policies.
Each task requires the robot to perform a specific action, such as picking up an object, opening a drawer, or stacking blocks. 
We follow the simulation setup of PerAct~\cite{shridhar2023perceiver} and RVT~\cite{goyal2023rvt} and use CopppelaSim~\cite{rohmer2013v} to simulate 18 RLBench tasks.
A Franka Panda robot with a parallel gripper is controlled to complete the tasks.
The 18 RLBench tasks are the same as PerAct and RVT. The visual observations are captured from four noiseless RGB-D cameras positioned at the front, left shoulder, right shoulder, and wrist with a resolution of 128×128.

\mypar{Baselines.}
We compare 3D-MVP with the following baselines on RLBench:

\noindent
(1) {\bf Image-BC}~\cite{jang2022bc} is an image-to-action behavior cloning approach which takes the visual observation and predict the corresponding action.
We compare with two variants which use CNN and ViT as the visual encoders, and call them Image-BC (CNN) and Image-BC (ViT), respectively;

\noindent
(2) {\bf C2F-ARM-BC}~\cite{james2022coarse} is another behavior cloning approach which converts RGB-D obversations to multi-resolution voxels and predicts the next key-frame action.

\noindent
(3) {\bf PerAct}~\cite{shridhar2023perceiver}: a multi-task a Perceiver transformer for robotic manipulation. The inputs are point clouds with color features and PerAct uses a Perceiver IO network to compress them to a fixed dimension~\cite{jaegle2021perceiver}.

\noindent
(4)
{\bf RVT}~\cite{goyal2023rvt}: The same Robotic View Transformer architecture as 3D-MVP but trained from scratch on the downstream tasks.
We do not compare with 2D pretraining methods since they do not work well on RLBench~\cite{pumacay2024colosseum}.

\mypar{Metrics.}
We report the task success rate for each individual tasks, and the average success rate.

\mypar{Results.}
We show quantitative results on Table~\ref{tab:rlbench}.
For the average success rate, 3D-MVP outperforms existing state-of-the-art methods, PerAct~\cite{shridhar2023perceiver} and RVT~\cite{goyal2023rvt}.
It demonstrates the effectiveness of bringing visual pretraining for manipulation policy which has explicit 3D modeling.
We find the improvement of 3D-MVP mainly comes from tasks which has medium difficulty, such as ``insert peg'', ``put in cupboard'', and ``stack blocks''.
If the task is too hard and PerAct/RVT is not able to solve any of them, the pretraining does not help.
If the task is too easy and PerAct/RVT has already reached more than 90\% success rate, the pretraining has limited space of improvement.

\subsection{Results on COLOSSEUM}
\label{sec:colosseum}

After validating our proposed pretraining is helpful for manipulation,
we further evaluate its generalization ability and robustness to environmental variations.

\mypar{Benchmark.} COLOSSEUM~\cite{pumacay2024colosseum} 
is a benchmark for evaluating generalization for robotic manipulation.
It contains 20 different tasks such as hockey, empty dishwasher.
For each task, it has 12 environmental perturbations, including changes in color, texture, size of objects and backgrounds, and lightnings.
The objects which can be changed include Manipulation object (\texttt{MO}), Receiver Object (\texttt{RO}) and the table.
Results on COLOSSEUM have also shown strong corelation with real robot experiments.
Therefore, it is well-suited and comprehensive for evaluating the generalization ability of manipulation approaches with pretraining on COLOSSEUM benchmark~\cite{pumacay2024colosseum}.

\begin{figure}
    \centering
    \includegraphics[width=\linewidth]{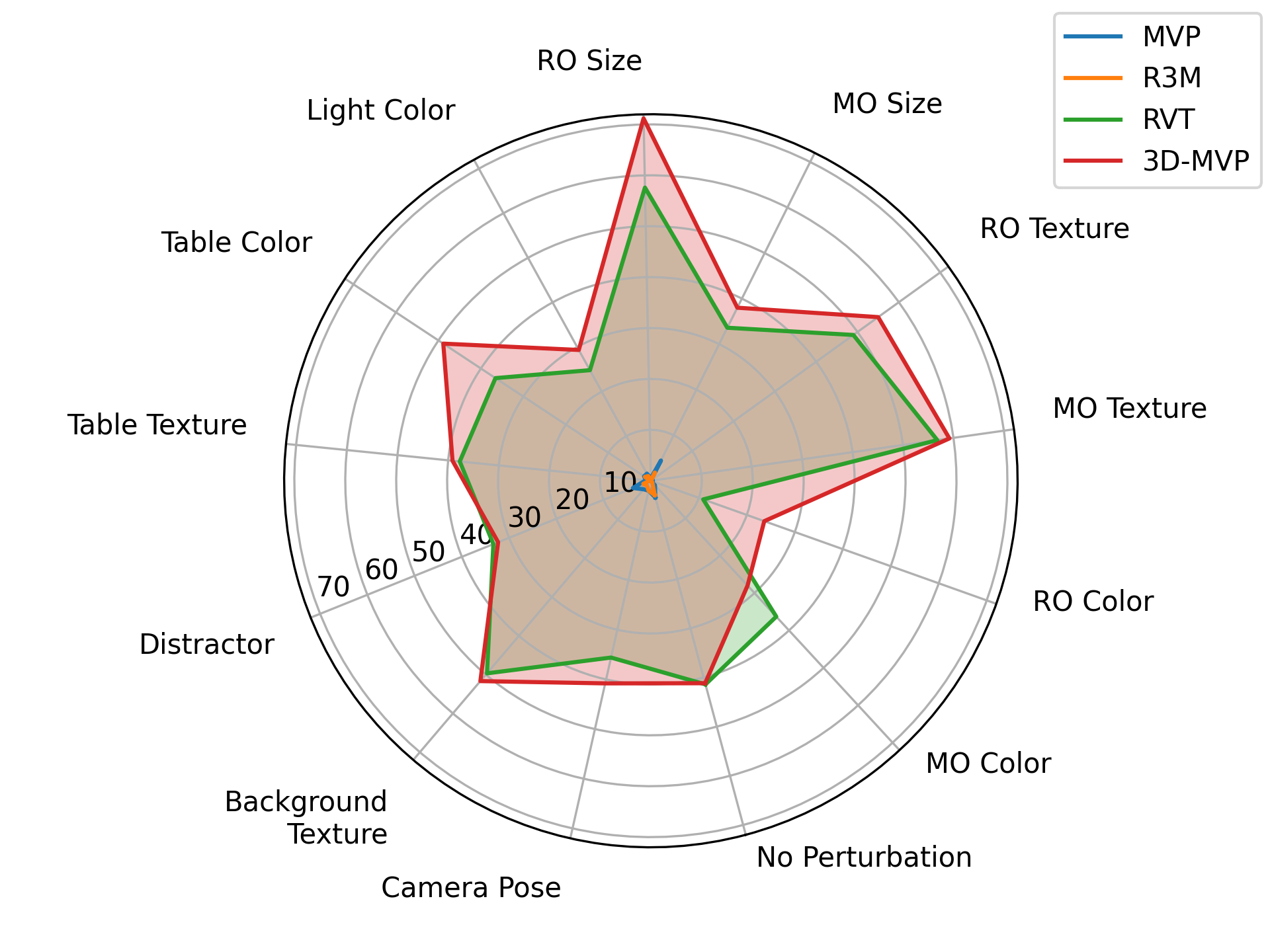}
    \caption{Results on COLOSSEUM~\cite{pumacay2024colosseum}. We report the average task completion success rate for 12 environmental perturbations and no perturbation. Manipulation policies which do explicit 3D reasoning (RVT~\cite{goyal2023rvt} works significantly better and 2D pretraining approaches (MVP~\cite{xiao2022masked} and R3M~\cite{nair2022r3m}). 3D-MVP is more robust than RVT on most perturbations. \texttt{MO} = manipulation object. \texttt{RO} = receiver object.}
    \label{fig:colosseum}
\end{figure}

\begin{table*}[t]
\centering
\begin{tabular}{lllc}
\toprule
Network Architecture & Pretraining Datasets & Masking Strategy & Success Rate \\
\midrule
3D-MVP & Objaverse (full) ~\cite{deitke2023objaverse} & RGB & 67.6 \\
3D-MVP & Objaverse (small)~\cite{deitke2023objaverse} & RGB & 65.3\\
3D-MVP & Objaverse (full)~\cite{deitke2023objaverse} & All & 64.4 \\
3D-MVP & 3D-FRONT~\cite{fu20213dfront} & RGB & 63.6\\
3D-MVP & RLBench~\cite{james2020rlbench} & RGB & 67.5\\
3D-MVP & RLBench~\cite{james2020rlbench} & All & 64.7\\
3D-MVP & None & None & 62.9\\
RVT~\cite{goyal2023rvt}  & None & None & 62.9\\
\bottomrule
\end{tabular}
\caption{Ablation studies on the RLBench benchmark. We analyze the contribution of our network architecture, pretraining datasets, and the masking stretegy. For each variant, we report the average task completion success rate on RLBench~\cite{james2020rlbench}.}
\label{tab:ablation}
\end{table*}

\mypar{Simulation setup.}
For simulation, we follow the original COLOSSEUM setup.
We use CoppelaSim~\cite{rohmer2013v} to simulate all tasks.
In training, we do not add any environmental perturbations and generate 100 demonstrations for each task.
During test time, we generate 12 environmental perturbations for each task.
For each environmental perturbation, we generate 25 demonstrations. 
For each demonstration, we repeat the generation 20 times if it fails.
In some cases, it is not possible to generate a plausible perturbation for some scenarios.
For example, we find it's hard to find the right perturbation parameters for the task of ``empty dishwasher''.
Therefore, we only report results on settings we are able to generate.
We report baselines on exactly the same setting to make sure the comparison is fair.

\mypar{Metrics.}
We also report the task completion success rate on COLOSSEUM.
Instead of averaging for each task, 
we report the average success rate for each environmental perturbation as it will highlight how each method is robust to different perturbations.

\mypar{Baselines.}
We compare 3D-MVP with state-of-the-art baselines reported on COLOSSEUM, which includes RVT and two 2D pretraining approaches:

\noindent
(1) MVP~\cite{xiao2022masked,radosavovic2023real}: 
a 2D pretraining approach using MAE reconstruction losses. It is pretrained on a collection of interaction datasets, such as Ego4D~\cite{grauman2022ego4d}, EpicKitchen~\cite{Damen18}, and 100DOH~\cite{Shan20}.
The pretrained encoder is then finetuned and evaluated on COLOSSEUM.

\noindent
(2) R3M~\cite{nair2022r3m}:
a 2D pretraining approach using a combination of reconstruction and contrastive losses. It is pretrained on Ego4D~\cite{grauman2022ego4d}. The pretrained encoder is then finetuned and evaluated on COLOSSEUM.

\noindent
(3)
RVT~\cite{goyal2023rvt}: Trained on COLOSSEUM from scratch.

\mypar{Results.}
We show results in Figure~\ref{fig:colosseum}.
First, our method outperforms existing 2D pretraining (MVP~\cite{xiao2022masked}, and R3M~\cite{nair2022r3m}) significantly.
It indicates existing 2D pretraining methods are not ready for complicated robotic manipulation.
Compared with RVT which is trained from scratch, our method is more robust to most perturbations.
It is especially robust to the change of texture and size of Receiver Object (\texttt{RO}), size of the manipulation object (\texttt{MO}), Light color and Table color.
We believe it is because the pretraining stage enables our approach to see diverse 3D objects.

\subsection{Ablation Studies}

To analyze the impact of different design choices in 3D-MVP, we conduct ablation studies on the RLBench benchmark. Table~\ref{tab:ablation} shows the average success rates of 3D-MVP with different network architecture, masking strategies and pretraining datasets.
And we discuss results as follows.

\mypar{Does the performance boost come from pretraining or network architecture?}
We first validate whether the performance boost comes from architecture change in Sec.~\ref{sec:pretraining} or the pretraining itself.
we finetune our model from scratch and find the performance is 62.9, similar to the original RVT~\cite{goyal2023rvt}.
It indicates the performance boost comes from our proposed 3D pretraining.

\mypar{Should we pretrain on object or room-level data?}
The choice of pretraining datasets are typically critical for self-supervised learning~\cite{dasari2023unbiased}. 
In our experiments, we mainly use Objaverse~\cite{deitke2023objaverse}, which is a object-centric 3D datasets.
Since we are mainly evaluated on tabletop manipulation, we also try room-level 3D datasets such as 3D-FRONT~\cite{fu20213dfront}.
We conduct the experiments on RLBench, and observe pretraining on 3D-FRONT only boosts the performance mildly from 62.9 to 63.6. In comparison, pretraining on Objaverse boosts the performance to 67.6.
We believe it is because 3D-FRONT has limited diversity and scale of objects and rooms compared with Objaverse.
And the diversity and scale of the pretraining data are one of the most important factors for learning generalizable representations.

\mypar{Does more pretraining data help?}
To validate it, we sampled 18K objects from Objaverse and called it Objaverse (small).
The full Objaverse dataset we use has 200K objects.
When we pretrain the encoder with Objaverse (small), we get a 65.3 mean success rate, which is worse than using Objaverse (full). This suggests that a larger dataset of pretraining helps performance in the downstream task.
If we can scale the training data to larger 3D model collections such as the full Objaverse-XL~\cite{deitke2024objaverse}, that might improve the performance further.

\begin{figure*}[t]
    \centering
    \includegraphics[width=\linewidth]{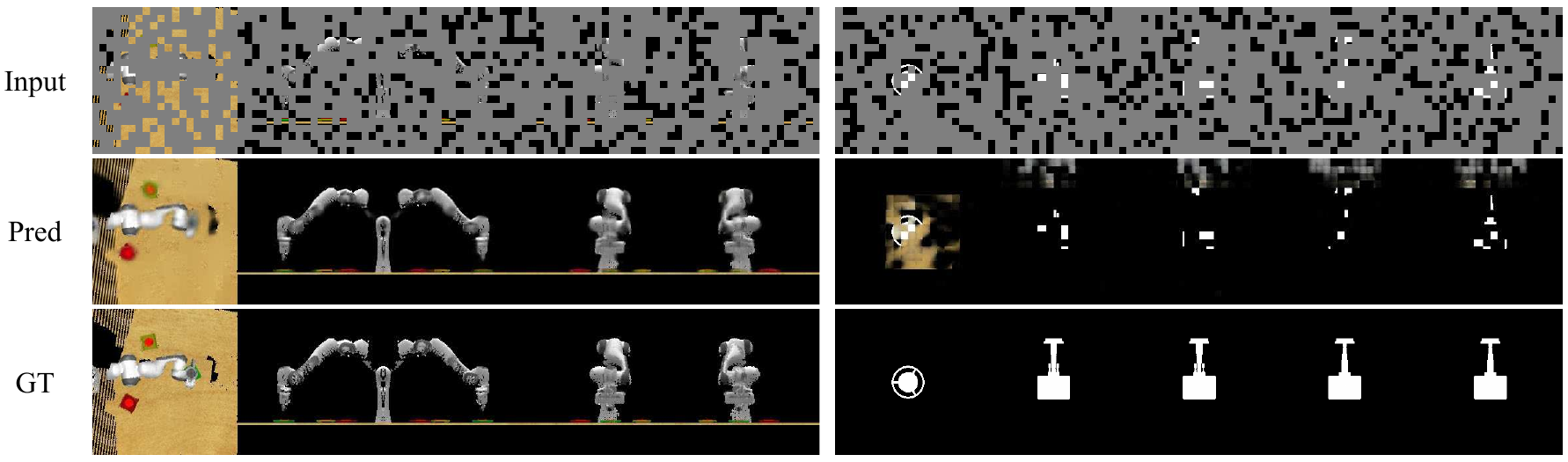}
    \caption{Pretraining MAE on RLBench scenes leads poor generalization performance. \textbf{(Left):} MAE reconstruction results on unseen RLBench renderings. \textbf{(Right)}: MAE reconstruction results on Objaverse renderings. While the reconstruction is reasonable on RLBench unseen renderings, it overfits to RLBench and does not learn a general representation.}
    \label{fig:rlbench_pretrain}
    \vspace{-1em}
\end{figure*}

\mypar{Can we pretrain on RLBench?}
Instead of relying on a different large-scale pretraining dataset such as Objaverse, we investigate whether we can just pretrain on RLBench point clouds, since the proposed 3D pretraining is self-supervised and only requires the 3D point cloud.
We extract the RLBench point cloud and build a pretraining dataset.
After the pretraining, we finetune the model on training demonstrations as usual.
We find it can achieve an average success rate of 67.5 on RLBench test set, which is comparable to Objaverse.
However, the pretrained model suffers from generalization issues.
As is shown in Figure~\ref{fig:rlbench_pretrain}, the encoder has overfitted to RLBench, and does not work on other environments such as COLOSSEUM.

\mypar{Masking strategy.}
We also observe that masking RGB channels performs better than masking all channels. We hypothesize that the network finds it very challenging to reconstruct all the channels and is unable to learn better visual representations. We view our findings as similar to He et al.~\cite{he2022masked}, who find that the benefits of pretraining diminish if the pretraining task becomes too difficult, like when the masking ratio becomes very high ($>$80\%).

\section{Conclusion}
\label{sec:conclusion}

In this work, we introduced 3D-MVP, a novel approach for 3D multi-view pretraining using masked autoencoders to improve the performance and generalization of robotic manipulation policies. By pretraining the encoder of Robotic View Transformer (RVT) on the large-scale Objaverse 3D object dataset using masked autoencoding, we demonstrate that the learned 3D representations lead to improved sample efficiency and robustness when finetuned on downstream manipulation tasks. We evaluated our approach on two benchmarks: RLBench, for general manipulation performance and COLOSSEUM, for systematic evaluation of robustness to environmental perturbations. On RLBench, 3D-MVP outperforms state-of-the-art manipulation baselines, achieving higher success rates. On COLOSSEUM, which tests 12 axes of variations such as object color, size, texture, lighting and more, 3D-MVP maintains higher success rates compared to baselines as the magnitude of perturbations increases. These results suggest that scalable 3D-aware pretraining on diverse object datasets is a promising approach to developing general-purpose robotic manipulation systems.

\mypar{Limitations and future work.}
While 3D-MVP achieves promising results on the RLBench and COLOSSEUM benchmarks, there are several limitations that we plan to address in future work.
First, the current version of 3D-MVP uses a fixed set of camera viewpoints and does not explicitly reason about occlusions and spatial relationships between objects. In future work, we plan to explore more advanced 3D representations, such as neural radiance fields, that can handle arbitrary camera viewpoints and model the 3D structure of the scene more explicitly.
Second, the current version of 3D-MVP assumes that the scene, robot, and objects follow quasi-static dynamics, and does not handle dynamic interactions between the robot and the environment. In future, we plan to explore techniques for learning action-conditional representations that can predict the effect of the robot's actions on the 3D scene.
Third, the current version of 3D-MVP requires a small amount of labeled data for each downstream task. In future, we plan to explore how to enable 3D-MVP to generate to novel manipulation tasks which have not been finetuned on.

\mypar{Social impacts.}
The development of more generalized and robust robotic manipulation systems enabled by 3D-MVP has the potential for significant societal impact. On the positive side, such systems could automate many repetitive or dangerous manual labor tasks, improving worker safety and productivity. Assisting robots could also improve quality of life for the elderly and people with disabilities. However, the increased automation from these systems may also displace some jobs, disproportionately impacting workers with lower levels of education and technical skills. Additionally, the datasets used for pretraining these models, like Objaverse, may encode biases that could be reflected in the downstream robotic system's behavior.

\mypar{Acknowledgment.}
We thank Ishika Singh and Jiafei Duan for their help with COLOSSEUM, and Jesse Zhang for his help with some baseline results.

{
    \small
    \bibliographystyle{ieeenat_fullname}
    \bibliography{local}
}

\end{document}